# Towards a Federated Learning Framework for Heterogeneous Devices of Internet of Things

Huanle Zhang
University of California, Davis
Davis, California
dtczhang@ucdavis.edu

Jeonghoon Kim
University of California, Davis
Davis, California
jhxkim@ucdavis.edu

## ABSTRACT

Federated Learning (FL) has received a significant amount of attention in the industry and research community due to its capability of keeping data on local devices. To aggregate the gradients of local models to train the global model, existing works require that the global model and the local models are the same. However, Internet of Things (IoT) devices are inherently diverse regarding computation speed and onboard memory. In this paper, we propose an FL framework targeting the heterogeneity of IoT devices. Specifically, local models are compressed from the global model, and the gradients of the compressed local models are used to update the global model. We conduct preliminary experiments to illustrate that our framework can facilitate the design of IoT-aware FL.

## 1 INTRODUCTION

The Internet of Things (IoT) has increasingly been deployed in various environments such as homes, hospitals, and industry, to help improve the convenience of life, energy efficiency, safety and security, and product quality [1]. With the increasing computational resources available, IoT devices can run sophisticated Machine Learning (ML) models [2]. For example, Amazon Echo converts human audio signals into a list of words, which are used for information searching.

In addition to running ML models, IoT devices can also train ML models. For example, Raspberry Pi 4 has four 1.5GHz CPU cores and 8GB memory at the price of tens of dollars. ML training on IoT devices has many benefits, including (1) *Privacy*. Data remain in local devices and avoid the data privacy issue, especially for healthcare applications; (2) *Adaptability*. An IoT device can continuously re-train its model while monitoring the environment. As a result, the model can adapt to new environments and cater to different human behaviors (i.e., model personalization).

Federated Learning (FL) has gained significant momentum in recent years, thanks to its ability to keep data in local devices while training an accurate global model [3]. Therefore, applying FL to the IoT domain has aroused a great interest in the community [4]. However, existing works require that the global and local models are the same during training [3], which omit that IoT devices are heterogeneous (speed, memory, etc.).

In this paper, we present a federated learning framework for IoT devices, especially targeting device heterogeneity. Figure 1 illustrates our system architecture. Our framework differs from existing works in that our local models are compressed from the global model, in order to meet the resource constraints of different devices. In other words, local models are compressed with different compression techniques (e.g., pruning and quantization) to different degrees (e.g., different pruning ratios). Consequently, existing gradient aggregation algorithms cannot be applied to our framework because the local models and the global models are not the same.

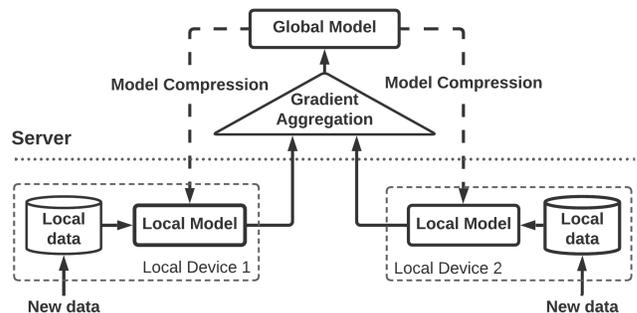

Figure 1: Our system architecture for federated learning. Local models are compressed from the global model, and their gradients are aggregated to retrain the global model, which in turn updates the local models.

To realize such a framework entails several challenges. First, current mainstream machine learning libraries, e.g., PyTorch and TensorFlow, do not support the training of the compressed models. To be flexible and extensible (e.g., quantify to different bit widths other than float and integer), we implement the whole system in C/C++ from scratch. Our framework includes a minimized but holistic model training pipeline such as forwarding and back prorogation of gradients. Our framework can facilitate the research on IoT-aware federated learning. Second, new gradient aggregation algorithms must exploit the gradients from the local compressed models to train the global model. Since local models can be compressed in different ways, it is difficult to design a gradient aggregation that can work for different compressed models, e.g., model 1 is pruned with a pruning ratio of 30%, and model 2 is quantified to an 8-bit integer. It could bring many potential research opportunities to design such gradient aggregation algorithms.

We conduct preliminary experiments to illustrate our framework. Specifically, we measure the accuracy, time overhead, and memory overheads when (1) different numbers of samples and (2) different data types (float64 and float32) are used for training. The results indicate that our framework can facilitate the research on FL for IoT devices.

## 2 MOTIVATION

To update the global model from the local models, existing methods require that the global model and the local models share the same model structure, so that the gradients from local models can be averaged to represent the gradient of the global model [3]. However, IoT devices vary in the capability of the computation speed and memory size. Due to the **device heterogeneity**, different IoT devices need to run different compressed versions of the global model. For example, an IoT hub can afford sophisticated models, whereas an embedded device can only run lightweight models. There is no existing work for training a global model by exploiting the gradients of local models that are differently compressed to the best of our knowledge.

Three compression techniques are widely used in the community [5]: (1) *Pruning*. It removes a model's weights that have a minor impact on the model's accuracy; (2) *Quantization*. It reduces the precision of the numbers (i.e., the bit width of the data) used to represent a model's weights; (3) *Clustering*. It groups the weights of each layer in a model into a predefined number of clusters, reducing the number of unique weight values. Therefore, our framework needs to support (1) the training of models compressed by pruning, quantization and clustering, and (2) the algorithms that can aggregate the gradients from differently compressed models to train the global model.

## 3 CHALLENGES

To realize such a framework for IoT-aware FL, we have the following main challenges.

### 3.1 A Platform for IoT-aware FL

Although commercial machine learning libraries exist, such as PyTorch, TensorFlow, and CNTK, none of them supports the training of compressed models. In our system architecture (Figure 1), the compressed models need to retrain themselves when new data are available and upload the gradients to the server for global model retraining. Please note that although PyTorch and TensorFlow support quantization-aware training [6], they only quantify the weights and outputs while still use float32 for calculations. Therefore, they do not represent the actual behavior of training quantified models. In addition, they do not support measurements of time and memory overheads of training a compressed model since they always conduct calculations in float32. Consequently, we need a new platform that is precise and flexible in training compressed models, gradient transmission from a local device to the server and local model update from the global model.

### 3.2 Algorithms for Aggregating Gradients of Local Models

The algorithms for aggregating gradients of local models that are differently compressed to train the global model are absent. This is partially due to the lack of a platform like ours to facilitate research on this topic. We hope that by using our platform, more research on gradient-based algorithms for FL will be accelerated. The absence of this kind of algorithms could offer many research opportunities for academia and powerful tools for real-world industry projects.

## 4 RELATED WORK

This section gives the related work regarding the knowledge transfer from the global model to the local models and from the local models to the global model.

### 4.1 Knowledge Transfer From Global Models to Local Models

In FL framework [3], the global and the local models share the same model structure, and their parameters are randomly initialized at first. At each round, the global model broadcasts the current model to the local clients on devices for the local training. After this step, local models are fine-tuned using local data and aggregated to update the global model. Local models on devices can also be personalized [7] by having additional personalization layers at the end of the model so that local models fine-tune parameters relying on their local private data. In [8], local models are pruned so that resource-limited devices can also train them.

### 4.2 Knowledge Transfer From Local Models to Global Models

Local models are often proprietary. Therefore, in addition to keeping the data in site, the owners also want to keep their local models in site. Furthermore, exposing local models [9] leads to security concerns, especially when the local models are used for the security domain (e.g., detecting malware);

Two main approaches to update global model without exposing local models [3]: (1) FedSGD, which updates gradients for the global model from those of local models, (2) FedAvg, which updates parameters of the global model by aggregating those of local models.

In FedSGD, each local client computes gradients of the model that the server broadcasts. The global model aggregates the averaged gradients of the local models trained on their local data, i.e., a single step of gradient descent is completed for each round. FedSGD may need a large number of communication rounds to converge. FedAvg can reduce the number of communication rounds by letting local models conduct multiple training rounds, and after training is completed over multiple local models, parameters, i.e., weights and bias in local models, are averaged out. Those averaged parameter values update the global model. Such a method is a practical way to implement the FL framework.

## 5 SYSTEM ARCHITECTURE

models are compressed from the global model using compression techniques such as pruning, quantization, and clustering. They receive more data during the model running on the local devices. For example, an attack detection model obtains more data from both benign traffic and malicious attacks. After receiving new data, the local model retrains itself using the new data and uploads the gradient to the server. On the server-side, the global model retrains itself by aggregating the gradients from local models. After the global model is retrained, the local model is updated by compressing the global model. The whole process repeats whenever new data is available to the local devices. As we can see, in our system architecture, local data remains within each local device, and only the gradients of the local models are exposed to the global model. Therefore, our framework keeps the merits of conventional federated learning, and meanwhile adapts to the device heterogeneity of IoT applications.

Compared to the server, which is usually equipped with a powerful CPU, GPU, and large memory, IoT devices are limited in computational resources. There are three essential metrics for local devices: the local model accuracy, the time overhead, and the memory overhead.

**Local Device Time Overhead.** As Equation (1) shows, the time overhead $T$ of the local device includes the time for local model training $T_{local}$, the time for gradient transmission from the local device to the server $T_{upload}$, the time for the global model training $T_{global}$, and the time for download the local model update $T_{download}$. Since IoT devices directly interact with humans, the local devices' time overhead should be small for a better user experience.

$$T = T_{local} + T_{upload} + T_{global} + T_{download} \quad (1)$$

With compressed local models, the local model training time $T_{local}$, the uploading time $T_{upload}$, and the downloading time $T_{download}$ are reduced compared to uncompressed models. Therefore, our framework has a shorter time overhead than traditional federated learning methods when the global model training time $T_{global}$ is the same.

**Local Device Memory Overhead.** Reducing memory overhead is crucial since IoT devices tend to have small memory. In addition, frequent memory access increases energy consumption, which is detrimental to the working hours of battery-powered IoT devices. Our framework has better memory overhead than traditional federated learning methods because the memory consumption of training a compressed model is less than the uncompressed model.

**Local Device Model Accuracy.** Although the accuracy of a compressed model is worse than its original model, the accuracy difference is not significant. For example, by quantifying a 32-bit float model to an 8-bit integer model, the accuracy is only reduced by 1.4% for a 100-layer ResNet model on ImageNet classification dataset [6]. However, in a federated learning scenario, the local model accuracy is affected by its local training and influenced by the global model accuracy. Therefore, it is too hasty to jump to the conclusion about local model accuracy change.

# 6 PRELIMINARY RESULTS

To illustrate that our framework could facilitate the research on IoT-aware FL, we conduct experiments to measure the model accuracy, the time overhead, and the memory overhead of training local models.

## 6.1 Experiment Setting

We implement a binary classification model using our framework and measure its performance in terms of accuracy, time, and memory overhead. Specifically, we build a 5-layer Multi-Layer Perception (MLP), where each layer has 10 neurons (filters) and uses sigmoid as the activation. To train and evaluate the model, we simulate data samples with 5 features using Gaussian distributions with the standard deviation of 1, where class 0 and class 1 has the mean value of -1 and 1, respectively. We use 1000 samples for validation and testing, respectively. We increase the number of training samples from 500 to 2000 to mimic the scenario that a local model witnesses more incoming data. We apply batch gradient descent for model training, which takes all the training data to update the model weights. All experiments are conducted in a Lenovo ThinkBook laptop (8 2.4GHz CPU cores and 16GB memory). Unless otherwise stated, all results are averaged by 20 runs.

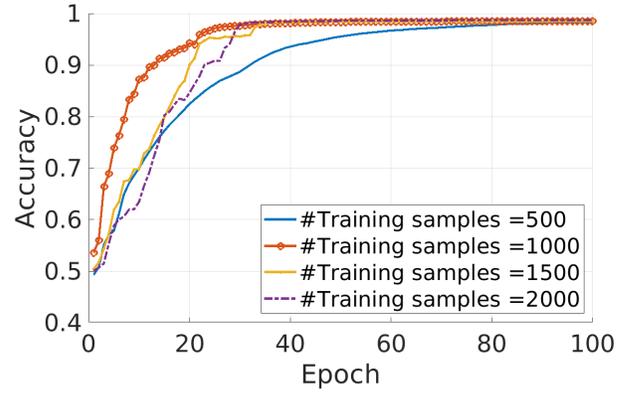

Figure 2: The model validation accuracy versus the the training epochs when different numbers of data samples are used for training.

## 6.2 Model Performance Versus Training Set Size

Local devices gradually receive more data, and thus the number of training samples is increasing. We investigate the behavior of the model training when a different number of training samples are used. We use float64 in this experiment.

Figure 2 shows the model validation accuracy versus the training epochs when a different number of training samples are used. As we can see, the model reaches the same maximum accuracy for a different number of training samples. When the training set is only 500 samples, it takes more epochs to achieve the maximum accuracy than larger training sets. For training sets that include more than 1000 samples, they require a similar number of epochs. However, the computation overheads are increased for larger training sets since each epoch in a larger batch size requires more time and memory.

Figure 3 shows the time and memory overhead for each epoch during model training when a different number of samples are used for training. The time and memory overheads for each epoch are linearly proportional to the sizes of the training sets. From Figure 2 and Figure 3(a), we can see that the training set of 1000 samples reaches the maximum accuracy faster than other sizes of training sets. Figure 3(b) shows the memory consumption for each epoch versus the training set sizes. As expected, smaller datasets have better memory footprints of model training. Our results indicate that different numbers of data samples for training can affect the system's training speed and memory requirement. It remains an open problem for deciding the optimal size of training samples, depending on the device hardware.

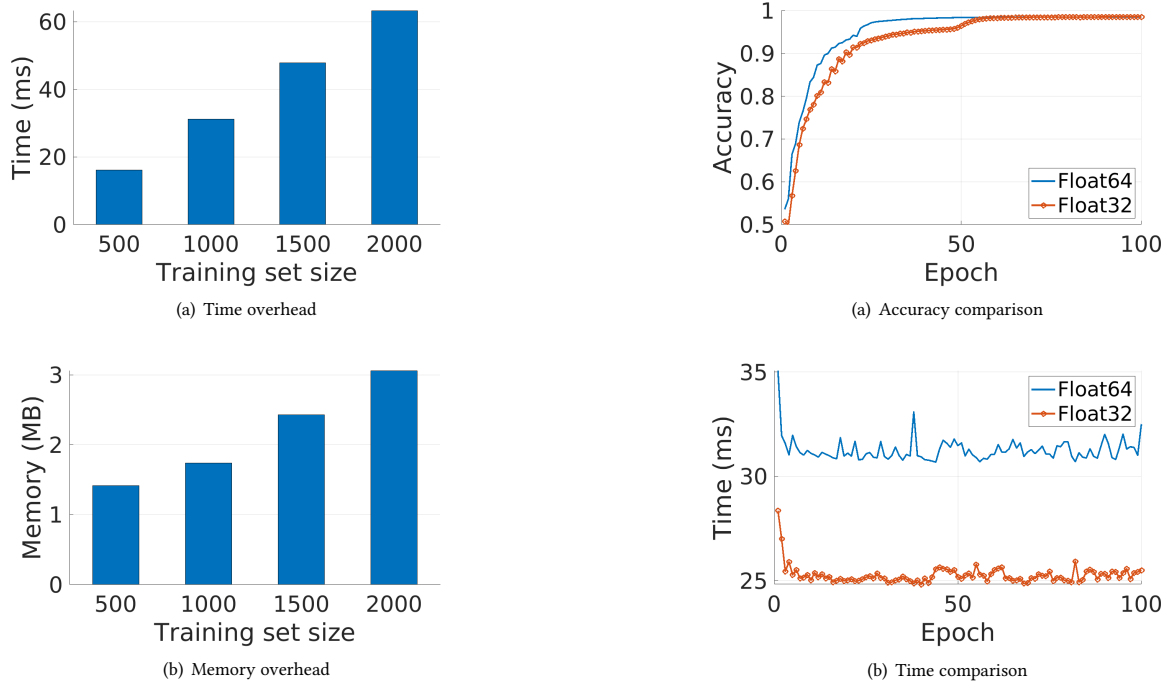

(a) Time overhead

(b) Memory overhead

**Figure 3: The time and memory overhead for each epoch during model training versus different sizes of training sets.**

### 6.3 Model Performance Versus Data Type

Different data types are expected to affect the training time, memory consumption, and model accuracy. We investigate the system performance when float32 and float64 are used to represent data. We use the training set of 1000 samples in this experiment.

Figure 4 compares the model accuracy, time overhead, and memory overhead between float64 and float32. We have the following observations: (1) Model Accuracy. Both data types result in the same maximum model accuracy. However, the larger data type requires fewer epochs to reach maximum accuracy. This is because when the data type has higher precision, the training process is more precise; (2) Time Overhead. Float64 is slower than float32 for training an epoch. Specifically, float64 and float32 take an average of 31.2ms and 25.3ms for each epoch, respectively. The spikes occur at the beginning (i.e., epoch 1) because it needs more time to initialize the models; (3) Memory Overhead. Float64 requires larger memory than float32. It needs an average of 1.7MB, while float32 only needs 1.2MB, with $\sim$ 30% memory reduction.

## 7 ONGOING WORK

We are still finishing the framework. Currently, our framework supports model training and measurement with several data types. We are actively working on the following aspects.

(a) Accuracy comparison

(b) Time comparison

(c) Memory comparison

**Figure 4: Accuracy, time, and memory comparison between float64 and float32 in model training.**

### 7.1 Supporting Data Types of Different Bit Width

Although 64 bits and 32 bits floating-points are most widely used for general-purpose calculations, other bit widths (e.g., 24 bits and 16 bits) are adopted by micro-processors. ML models run faster and take less memory when using data types of smaller bit width, with the compromise of precision loss. To fully understand model performance regarding quantization of different bit widths, we plan to implement various data types by adjusting the number of bits for the exponent and the significand of floating numbers, based on the IEEE standard for floating-point arithmetic [10]. In addition to floating points, integers are also widely used to reduce time and memory overheads.

## 7.2 Supporting Mainstream ML Models

Our current implementation only supports simple model structures such as MLP. Although it is not our priority to implement all ML model structures and layers, we plan to support mainstream models specifically designed for mobile devices, such as MobileNetV2 [11] and ShuffleNetV2 [12]. In addition, we are interested in supporting Three-Dimensional (3D) ML models, considering the emerging applications of 3D vision [13]. For example, Virtual Reality (VR) and Augmented Reality (AR) heavily rely on 3D ML models to realize fascinating user applications.

## 7.3 Designing Algorithms for Aggregating Gradients of Local Models

Last but not least, we plan to leverage our framework to design algorithms that can aggregate gradients of compressed models to train the global model. It brings both the research challenges and opportunities, in light of there is a variety of compression techniques and ratios. For example, one can design a gradient aggregation algorithm for different pruning ratios, while another one can design an algorithm for models that are compressed using pruning and quantization at the same time.

## 8 DISCUSSION AND CONCLUSION

In this paper, we propose an FL framework that targets heterogeneous IoT devices. Our preliminary experiments show some promising outputs to illustrate that our FL framework can facilitate the design of IoT-aware FL.

We first apply our framework to investigate how the model accuracy, time overhead, and memory overhead change against different sizes of training sets. This simple experiment sheds light on the importance of the training set size for IoT devices, which the community has omitted. Then, we explore the model performance when different data types are used for model training. The results are analogous to the performance comparison between different quantization ratios.

In future research, other compression techniques such as pruning and clustering can be potentially applied, and also, mainstream models designed to support mobile devices should be involved. Furthermore, communication between the server and the local devices, particularly updating the global model part, is not fully addressed yet, so that this critical part should take into consideration.